\documentclass[11pt,a4paper]{article}

\usepackage[hyperref]{acl2021}

\usepackage{times}
\usepackage{latexsym}

\usepackage[T1]{fontenc}

\usepackage[utf8]{inputenc}

\usepackage{microtype}
\usepackage{times}
\usepackage{latexsym}
\usepackage{lipsum}
\usepackage{booktabs}
\usepackage{array}
\usepackage{multirow}
\usepackage{graphicx}
\usepackage[normalem]{ulem}

\usepackage{etoolbox}
\usepackage{todonotes}
\usepackage{hyperref}
\usepackage{enumerate}
\usepackage{enumitem}
\usepackage{amsmath}
\usepackage{color}
\usepackage{tcolorbox}
\usepackage{CJK}
\usepackage{adjustbox}
\tcbset{width=0.9\textwidth,boxrule=0pt,colback=red,arc=0pt,auto outer arc,left=0pt,right=0pt,boxsep=5pt}
\usepackage[font=small,skip=-5pt]{subcaption}
\usepackage{diagbox}

\definecolor{citeworthy}{HTML}{32A852}

\usepackage{microtype}

\aclfinalcopy 

\newcommand\modelname{\textsc{CiteBert}}
\newcommand\dataset{\textsc{CiteWorth}}

\title{\dataset: Cite-Worthiness Detection for Improved Scientific Document Understanding}

\author{Dustin Wright \and Isabelle Augenstein \\
  Dept. of Computer Science \\
  University of Copenhagen \\
  Denmark \\
  \texttt{\{dw|augenstein\}@di.ku.dk}}

\date{}

\begin{document}
\maketitle
\begin{abstract}
Scientific document understanding is challenging as the data is highly domain specific and diverse. However, datasets for tasks with scientific text require expensive manual annotation and tend to be small and limited to only one or a few fields. At the same time, scientific documents contain many potential training signals, such as citations, which can be used to build large labelled datasets. Given this, we present an in-depth study of cite-worthiness detection in English, where a sentence is labelled for whether or not it cites an external source. To accomplish this, we introduce \dataset, a large, contextualized, rigorously cleaned labelled dataset for cite-worthiness detection built from a massive corpus of extracted plain-text scientific documents. We show that \dataset~is high-quality, challenging, and suitable for studying problems such as domain adaptation. Our best performing cite-worthiness detection model is a paragraph-level contextualized sentence labelling model based on Longformer, exhibiting a 5 F1 point improvement over SciBERT which considers only individual sentences. Finally, we demonstrate that language model fine-tuning with cite-worthiness as a secondary task leads to improved performance on downstream scientific document understanding tasks.

\end{abstract}

\section{Introduction}
Building effective NLP systems from scientific text is challenging due to the highly domain-specific and diverse nature of scientific language, and a lack of abundant sources of labelled data to capture this. While large scale repositories of extracted, structured, and unlabelled plain-text scientific documents have recently been introduced~\cite{lo2020s2orc}, most datasets for downstream tasks such as named entity recognition~\cite{li2016biocreative} and citation intent classification~\cite{cohan2019structural} remain limited in size and highly domain specific. This begs the question: what useful training signals can be automatically extracted from massive unlabelled scientific text corpora to help improve systems for scientific document processing?

Scientific documents contain much inherent structure (sections, tables, equations, citations, etc.), which can facilitate creating large labelled datasets. Some recent examples 
include using paper field~\cite{beltagy2019scibert}, the section to which a sentence belongs~\cite{cohan2019structural}, and the cite-worthiness of a sentence~\cite{cohan2019structural,sugiyama2010identifying} as a training signal.

Cite-worthiness detection is the task of identifying \textit{citing sentences}, i.e. sentences which contain a reference to an external source. It has useful applications, such as in assistive document editing, and as a first step in citation recommendation~\cite{farber2018cite}. In addition, cite-worthiness has been shown to be useful in helping to improve the ability of models to learn other tasks~\cite{cohan2019structural}. 
We also hypothesize that there is a strong domain shift between how different fields use citations, and that such a dataset is useful for studying domain adaptation problems with scientific text.


However, constructing such a dataset to be of high quality is surprisingly non-trivial. Building a dataset for cite-worthiness detection involves extracting sentences from a scientific document, labelling whether each sentence contains a citation, and removing all citation markers. As a form of distant supervision, this naturally comes with the hazard of adding spurious correlations, such as poorly removed citation text causing ungrammatical sentences and hanging punctuation, which can trivially indicate a cite-worthy or non-cite-worthy sentence. Additionally, the task itself is quite difficult to learn, as different fields employ citations differently, and whether or not a sentence contains a citation depends on factors such as the context in which it appears. Given this, we present \dataset, a 
rigorously curated dataset for cite-worthiness detection in English. \dataset~contains rich metadata, such as authors and links to cited papers, and all data is provided in \textit{full paragraphs}: every sentence in a paragraph is labelled in order to provide sentence \textit{context}. We offer the dataset to the research community to facilitate further research on cite-worthiness detection and related scientific document processing tasks.


Using \dataset, we ask the following primary research questions:
\begin{quote}
\textbf{RQ1}: How can a dataset for cite-worthiness detection be automatically curated with low noise (\S\ref{sec:dataset-construction})?

\textbf{RQ2}: What methods are most effective for automatically detecting cite-worthy sentences (\S\ref{sec:benchmarks})?

\textbf{RQ3}: How does domain affect learning cite-worthiness detection (\S\ref{sec:domain-adaptation})?

\textbf{RQ4}: Can large scale cite-worthiness data be used to perform transfer learning to downstream scientific text tasks (\S\ref{sec:transfer-learning})?
\end{quote}
We demonstrate that \dataset~is of high quality through a manual evaluation, that there are large differences in how models generalize to data from different fields, and that sentence context leads to significant performance improvements on cite-worthiness detection. Additionally, we find that cite-worthiness is a useful task for transferring to downstream scientific text tasks, in particular citation intent classification, for which we offer performance improvements over the current state-of-the-art model SciBERT~\cite{beltagy2019scibert}.


In sum, our \textbf{contributions} are as follows:
\begin{itemize}[noitemsep]
    \item \dataset, a dataset of 1.2M rigorously cleaned sentences from scientific papers labelled for cite-worthiness, balanced across 10 diverse scientific fields.
    \item A method for cite-worthiness detection which considers the entire paragraph a sentence resides in, improving by 5 F1 points over the state of the art model for scientific document processing, SciBERT~\cite{beltagy2019scibert}.
    \item A thorough analysis of the problem of cite-worthiness detection, including explanations of predictions and insight into how scientific domain affects performance.
    \item New state of the art on citation intent detection via transfer learning from joint citation detection and language model fine-tuning on \dataset, with improved performance over SciBERT on several other tasks.
\end{itemize}
\section{Related Work}
\subsection{Cite-Worthiness Detection}
Cite-worthiness detection is the task of identifying \textit{citing sentences}, i.e. sentences which contain a reference to an external source. The reasons for citing are varied, e.g. to give credit to existing ideas or to provide evidence for a claim being made. 
\citet{sugiyama2010identifying} perform cite-worthiness detection using SVMs with features such as unigrams, bigrams, presence of proper nouns, and the classification of previous and next sentences. They create a dataset from the ACL Anthology Reference corpus (ACL-ARC, \citet{bird2008acl}), using heuristics to remove citation markers. \citet{farber2018cite} document the performance of convolutional recurrent neural nets on a larger set of three datasets coming from ACL-ARC, arXiv CS~\cite{farber2018high}, and Scholarly Dataset 2.\footnote{\url{http://www.comp.nus.edu.sg/~sugiyama/SchPaperRecData.html}} Datasets from these studies suffer from high class imbalance, are limited to only one or a few domains, and little analysis of the datasets is performed to understand the quality of the data or what aspects of the problem are difficult or easy. Additionally, no study to date has considered how sentence context can affect learning to perform cite-worthiness detection.

In addition to being a useful task in itself, 
cite-worthiness detection is useful for other tasks in scientific document understanding. In particular, it has been shown to help improve performance on the closely related task of citation intent classification~\cite{jurgens2018measuring} when used as an auxiliary task in a multi-task setup~\cite{cohan2019structural}. However, cite-worthiness detection has not been studied in a transfer learning setup as a pretraining task for multiple scientific text problems. In this work, we seek to understand to what extent cite-worthiness detection is a transferable task.

\paragraph{Scientific Document Understanding}
Numerous problems related to scientific document understanding have been studied previously. Popular tasks include named entity recognition~\cite{li2016biocreative,kim2004introduction,dougan2014ncbi,luan-etal-2018-multi} and linking~\cite{conf/akbc/WrightKMH19}, keyphrase extraction~\cite{augenstein2017semeval,augenstein-sogaard-2017-multi}, relation extraction~\cite{kringelum2016chemprot,luan-etal-2018-multi}, dependency parsing~\cite{kim2003genia}, citation prediction~\cite{holm2020longitudinal}, citation intent classification~\cite{jurgens2018measuring,cohan2019structural}, summarization~\cite{collins-etal-2017-supervised}, and fact checking~\cite{Wadden2020FactOF}.

Datasets for scientific document understanding tasks tend to be limited in size and restricted to only one or a few fields, making it difficult to build models with which one can study cross-domain performance and domain adaptation. Here, we curate a large dataset of cite-worthy sentences spanning 10 different fields, showing that such data is both useful for studying domain adaptation and for transferring to related downstream scientific document understanding tasks.

\section{RQ1: \dataset~Dataset Construction}
\label{sec:dataset-construction}
\begin{table*}[t]
    \centering
    \fontsize{10}{10}\selectfont
    \begin{tabular}{p{15cm}}
    \toprule 
    \textbf{Biology} \\
    \midrule
    Wood Frogs (Rana sylvatica) are a charismatic species of frog common in much of North America. They breed in explosive choruses over a few nights in late winter to early spring. \textcolor{citeworthy}{\emph{The incidence in Wood Frogs was associated with a die-off of frogs during the breeding chorus in the Sylamore District of the Ozark National Forest in Arkansas\sout{ (Trauth et al., 2000)}.}} \\
    \bottomrule \\
    \toprule
    \textbf{Computer Science} \\
    \midrule
    \textcolor{citeworthy}{\emph{Land use or cover change is a direct reflection of human activity, such as land use, urban expansion, and architectural planning, on the earth's surface caused by urbanization\sout{ [1].}}} Remote sensing images are important data sources that can efficiently detect land changes. \textcolor{citeworthy}{\emph{Meanwhile, remote sensing image-based change detection is the change identification of surficial objects or        geographic phenomena through the remote observation of two or more different phases\sout{ [2]}.}} \\
    \bottomrule 

    \end{tabular}
    \caption{Excerpts from training samples in \dataset~from the Biology and Computer Science fields. Green sentences are cite-worthy sentences, from which citation markers are removed during dataset construction.}
    \label{tab:dataset_examples}
\end{table*}

The first research question we ask is: How can a dataset for cite-worthiness detection be automatically curated with low noise? To answer this, we start with the S2ORC dataset of extracted plain-text scientific articles~\cite{lo2020s2orc}. It consists of data from 81.1M English scientific articles, with full structured text for 8.1M articles. S2ORC uses \textsc{ScienceParse}\footnote{\url{https://github.com/allenai/scienceparse}} to parse PDF documents and \textsc{Grobid}\footnote{\url{https://github.com/kermitt2/grobid}} to extract structured data from text. As such, the data also includes rich metadata, e.g. Microsoft Academic Graph (MAG) categories, linked citations, and linked figures and tables. Throughout this work, a ``citation span'' denotes a span containing citation text (e.g. ``[2]''), and a ``citation marker'' is any text that trivially indicates a citation, such as the phrase ``is shown in.'' A citation span is also a type of citation marker. It is important to remove all citation markers from the dataset to prevent the model learning to use these signals for prediction.

\subsection{Data Filtering}
\label{sec:data-filtering}
Given the size of S2ORC, we first reduce the candidate set of data to papers where all of the following are available. 
\begin{itemize}[noitemsep]
    \item Abstract
    \item Body text
    \item Bibliography
    \item Tables and figures
    \item Venue information
    \item Inbound citations
    \item Microsoft Academic Graph categories
\end{itemize}
Filtering based on these criteria results in 5,494,387 candidate papers from which to construct the dataset.
After filtering the candidate set of papers, we perform the following checks on the sentences in the body text.
\begin{enumerate}[noitemsep]
    \item Citation spans are parenthetical author-year or bracketed-numerical form.
    \item Citation spans are at the end of a sentence.
    \item All possible citation spans have been extracted by S2ORC. 
    \item No citation markers are left behind after removing citation spans from the text.
    \item Sentence starts with a capital letter, ends with `.', `!', or `?', and is at least 20 characters long.
\end{enumerate}
The detailed steps of extracting and labelling sentences based on these criteria are given in \S\ref{sec:extraction}. With the first two criteria, we restrict the scope of cite-worthy sentences to being only those whose citation span comes at the end of a sentence, and whose citation format is parenthetical author-year form or bracketed-numerical form. In other words, cite-worthy sentences in our data are constrained to those of the following forms.

\begin{quote}
    This result has been shown in previous work (Author1 et al., \#\#\#\#, ...).
    
    This result has been shown in previous work [\#-\#].
\end{quote}
In this, we ignore citation sentences which contain inline citations, such as ``The work of Authors et al. (\#\#\#\#) has shown this in previous work'', as well as any sentence with a citation format that does not match the two we have selected. 

Curating cite-worthy sentences as such helps prevent spurious correlations in the data. Removing citations in the middle of a sentence runs the risk of rendering the sentence ungrammatical (for example, the above sample would turn into ``The work \textit{of has} shown this in previous work''), providing a signal to machine learning models. While there are cases where inline citations could potentially be removed in their entirety and not destroy the sentence structure, this is beyond the scope of this paper and left to future work.

\subsection{Extracting Cite-Worthy Sentences in Context}
\label{sec:extraction}
As we are interested in using sentence context for prediction, we perform extraction at the \textit{paragraph level}, ensuring that all of the sentences in a given paragraph meet the checks given in \S\ref{sec:data-filtering}. As such, our dataset construction pipeline for a given paper begins by first extracting all paragraphs from the body text which belong to sections with titles coming from a constrained list of permissible titles (e.g. ``Introduction,'' ``Methods,'' ``Discussion'') . The full list is provided in \autoref{sec:premissible-section-titles}. 

For a given paragraph, we first word and sentence tokenize the text with SciSpacy~\cite{neumann-etal-2019-scispacy}. Each sentence is then checked for containing citations using the provided citation spans in the S2ORC dataset. In some cases, the sentence contains citations which were missed by S2ORC; these are checked using regular expressions (see \autoref{sec:regexes}). If a match is found the paragraph is ignored, as we only consider paragraphs where all citations have been extracted by S2ORC. Otherwise, the location and format of the citation is checked, again using regular expressions (see \autoref{sec:regexes}). If the citation is not at the end of the sentence, the paragraph is ignored. We then remove the citation text using the provided citation spans for all sentences which pass the above checks.

Simply removing the citation span runs the risk of leaving other types of citation markers, such as hanging punctuation and prepositional phrases e.g. ``This was shown by the work of \sout{Author et al. (\#\#\#\#)}.'' To mitigate this, we remove all hanging punctuation at the end of a sentence that is not a period, exclamation point, or question mark, and check for possible hanging citations using the regular expression provided in \autoref{sec:regexes}. The regular expression checks for many common prepositional phrases and citation markers occurring as the last phrase of a sentence such as ``see,'' ``of,'' ``by,'' etc. 

To handle issues with sentence tokenization, we also ensure that the first character of each sentence is a capital letter, and that the sentence ends with a period, exclamation point, or question mark. If all criteria are met for all sentences in a paragraph, the paragraph is added to the dataset. Finally, we build a dataset which is diverse across domains by evenly sampling paragraphs from the following 10 MAG categories, ensuring that each paragraph belongs to exactly one category: Biology, Medicine, Engineering, Chemistry, Psychology, Computer Science, Materials Science, Economics, Mathematics, and Physics. Example excerpts from the dataset are presented in \autoref{tab:dataset_examples}, and the statistics for the final dataset are given  in~\autoref{tab:dataset_statistics}.\footnote{The full dataset can be downloaded from this repository: \url{https://github.com/copenlu/cite-worth}}

\begin{table}[t]
    \centering
    \fontsize{10}{10}\selectfont
    \begin{tabular}{l r}
    \toprule 
    Metric & \# \\
    \midrule
    Total sentences & 1,181,793 \\
    Total number of tokens & 34,170,708 \\
    Train sentences & 945,426 \\
    Dev sentences & 118,182 \\
    Test sentences & 118,185 \\
    Total cite-worthy & 375,388 (31.76\%)\\
    Total non-cite-worthy & 806,405 (68.24\%)\\
    Min char length & 21 \\
    Max char length & 1,447 \\
    Average char length & 152 \\
    Median char length & 142 \\
    \bottomrule 

    \end{tabular}
    \caption{Various statistics of the \dataset~dataset.}
    \label{tab:dataset_statistics}
    \vspace{-4mm}
\end{table}

\subsection{Manual Evaluation}
In order to provide some measure of the general quality of \dataset, we perform a manual evaluation of a sample of the data. We annotate the data for whether or not citation markers are completely removed, and for whether or not the sentences are well-formed, containing no obvious extraction artifacts. We sample 500 cite-worthy sentences and 500 non-cite-worthy sentences randomly from the data. Additionally, we compare to a baseline where the only heuristic used is to remove citation spans based on the provided spans in the S2ORC dataset. We again sample 500 cite-worthy and 500 non-cite-worthy sentences for annotation. The two sets are shuffled together and given to an independent expert annotator with a PhD in computer science for labelling. The annotator is instructed to label if the sentences are complete and have no hanging punctuation or obvious extraction errors, and if there are any textual indicators that the sentences contain a citation. The results for the manual annotation can be seen in \autoref{tab:manual_annotation_results}.

\begin{table}[t]
    \centering
    \fontsize{10}{10}\selectfont
    \begin{tabular}{l c c}
    \toprule 
    Method & Extracted Correct & Markers Removed \\
    \midrule
    Baseline & 92.07 & 92.78 \\
    Ours & \textbf{98.90} & \textbf{98.10} \\
    \bottomrule 

    \end{tabular}
    \caption{Results of manually annotating 1000 random sentences (per method) from \dataset~and a naive baseline which only removes citations based on provided citation spans . ``Extracted Correct'' are results for correctly extracting the sentences (i.e. that sentences are tokenized correctly and are grammatical), and ``Markers Removed'' are results for successfully removing citation markers. The data curated using our method has ~6\% fewer errors in terms of extraction and removal of citation markers, and less than 2\% of the samples have some form of citation marker.}
    \label{tab:manual_annotation_results}
\end{table}

We see that the \dataset~data are of a much higher quality than removing citation markers based only on the citation spans. Overall, our heuristics improve on extraction quality by 6.83\% absolute and on removing markers of citations by 5.32\% absolute. This results in 1.1\% of the sample data containing sentence cleaning issues, and 1.9\% having trivial markers indicating a citation is present. 
We argue that this is a strong indicator of the quality of the data for supervised learning.

  

  

\section{RQ2: System Evaluation\footnote{The code for all experiments can be found here: \url{https://github.com/copenlu/cite-worth}}}
\label{sec:benchmarks}

Next, we ask: what methods are most effective for performing cite-worthiness detection? To answer this and characterize the difficulty of the problem, we run a variety of baseline models on \dataset. The hyperparameters selected for each model, as well as hyperparameter sweep information, are given in Appendix \ref{sec:hyperparams}.

\paragraph{Logistic Regression}
As a simple baseline, we use a logistic regression model with TF-IDF input features. 

\paragraph{\citet{farber2018cite}} The convolutional recurrent neural network (CRNN) model from \citet{farber2018cite}. They additionally use oversampling to deal with class imbalance.

\paragraph{Transformer} We additionally train a Transformer model from scratch~\cite{vaswani2017attention}, tuning the model hyperparameters on a subset of the training data via randomized grid search.

\paragraph{BERT} We use a pretrained BERT model~\cite{devlin2019bert} due to the strong performance of large pretrained Transformer models on downstream tasks.

\paragraph{SciBERT} SciBERT~\cite{beltagy2019scibert} is a BERT model pretrained on a large corpus of scientific text from Semantic Scholar~\cite{ammar2018construction}, and is therefore potentially better suited to fine-tuning on scientific cite-worthiness detection.

\paragraph{SciBERT + PU Learning} We experiment with SciBERT trained using positive-unlabelled (PU) learning~\cite{elkan2008learning} which has been shown to significantly improve performance on citation needed detection in Wikipedia and rumour detection on Twitter~\cite{wright2020fact}. The intuition behind PU learning is to assume that cite-worthy data is labelled and non-cite-worthy data is unlabelled, containing some cite-worthy examples. This is to mitigate the subjectivity involved in adding citations to sentences. Technically, this involves training a classifier on the positive-unlabeled data which will predict the probability that a sample is labeled, and using this to estimate the probability that a sample is positive given that it is unlabeled. One then trains a second model where positive samples are trained on normally and unlabeled samples are duplicated and trained on twice, once as positive and once as negative data, weighed by the first model's estimate of the probability that the sample is positive.

  

\paragraph{Longformer-Ctx} Finally, we test our novel contextualized prediction model based on Longformer~\cite{DBLP:journals/corr/abs-2004-05150}. Longformer is a Transformer based language model which uses a sparse attention mechanism to scale better to longer documents. 
We process an entire paragraph at a time, separating each sentence with a \texttt{[SEP]} token. Each \texttt{[SEP]} token representation at the output of Longformer is then passed through a network with one hidden layer and a classifier. As a control, we also experiment with Longformer using only single sentences as input (Longformer-Solo).

    
    

Due to the imbalance in the distribution of classes, the loss for each of the models is weighted. For comparison, we include results for SciBERT without weighting the loss function. The results for our baseline models on the test set of the dataset are given in \autoref{tab:citation_detection_results}.

\begin{table}[t]
    \setlength{\tabcolsep}{1.5pt}
    \def\arraystretch{1.2}
    \centering
    \fontsize{10}{10}\selectfont
    \begin{tabular}{l c c c}
    \toprule 
    Method & P & R & \multicolumn{1}{c}{F1} \\
    \midrule 
       Logistic Regression & $46.65_{0.00}$& $64.88_{0.00}$& $54.28_{0.00}$\\
       \citet{farber2018cite} & $49.57_{0.96}$& $65.56_{2.61}$& $56.41_{0.34}$\\
       Transformer & $47.92_{0.78}$& $71.59_{1.74}$& $57.39_{0.10}$\\
       BERT & $55.04_{0.66}$& $69.02_{1.33}$& $61.23_{0.21}$\\
       SciBERT-no-weight & $\mathbf{65.94}_{0.37}$& $51.62_{0.53}$& $57.91_{0.30}$\\
       SciBERT & $57.03_{0.50}$& $68.08_{1.03}$& $62.06_{0.15}$\\
       SciBERT + PU & $49.46_{0.83}$& $\mathbf{82.12}_{1.40}$& $61.73_{0.27}$\\
       Longformer-Solo & $57.21_{0.25}$& $68.00_{0.41}$& $62.14_{0.02}$\\
       Longformer-Ctx & $59.92_{0.28}$& $77.15_{0.49}$& $\mathbf{67.45}_{0.06}$\\
    \bottomrule 

    \end{tabular}
    \caption{F1 performance of baselines on the test set of \dataset. Results are averaged across 5 seeds, with standard deviations given in the subscripts.}
    \label{tab:citation_detection_results}
\end{table}
Our results indicate that context is critical, resulting in the best F1 score of 67.45 (Longformer-Ctx) and a 5.31 point improvement over the next best model. Using class weighting is also highly important, resulting in another increase of over 4 F1 points. Compared to not using class weights, PU learning performs significantly better, and leads to the highest recall of all models under test. Additionally, language model pre-training is useful, as BERT, SciBERT, and Longformer all perform significantly better than a Transformer trained from scratch and the model from \citet{farber2018cite}. 

To gain some insight into what the model learns, we visualize the most salient features from SciBERT for selected easy and hard examples. We use the single-sentence model instead of the paragraph model for simplicity. ``Easy'' samples are defined as those which the model predicted correctly with high confidence, and ``hard'' examples are defined as those for which the model had low confidence in its prediction. We use the InputXGradient method~\cite{kindermans2016investigating}, specifically the variant using L2 normalization over neurons to get a pre-embedding score, as it has been recently shown to have the best overall agreement with human rationales versus several other explainability techniques~\cite{atanasova2020diagnostic}. The method works by calculating the gradient of the output with respect to the input, then multiplies this with the input. In the examples below ``C'' refers to an example whose gold label is cite-worthy, and ``N'' refers to an example whose gold label is non-cite-worthy.

The model is able to pick up on obvious markers of cite-worthy and non-cite-worthy sentences for the following correctly classified examples, such as that a sentence refers to a preprint or to different sections within the paper itself:
\begin{quote}
C: 
\begin{CJK}{UTF8}{gbsn}
{\setlength{\fboxsep}{0pt}\colorbox{white!0}{\parbox{0.35\textwidth}{
\colorbox{red!12.105263157894738}{\strut [CLS]} \colorbox{red!4.7894736842105265}{\strut in} \colorbox{red!5.7894736842105265}{\strut this} \colorbox{red!13.157894736842106}{\strut note} \colorbox{red!3.736842105263158}{\strut ,} \colorbox{red!8.947368421052634}{\strut we} \colorbox{red!10.526315789473685}{\strut follow} \colorbox{red!5.052631578947369}{\strut the} \colorbox{red!9.473684210526317}{\strut approach} \colorbox{red!5.7894736842105265}{\strut to} \colorbox{red!5.105263157894737}{\strut the} \colorbox{red!8.421052631578949}{\strut en} \colorbox{red!20.526315789473685}{\strut \#\#och} \colorbox{red!7.894736842105264}{\strut \#\#s} \colorbox{red!23.157894736842106}{\strut conjecture} \colorbox{red!22.105263157894736}{\strut outlined} \colorbox{red!10.526315789473685}{\strut in} \colorbox{red!12.105263157894738}{\strut the} \colorbox{red!100.0}{\strut preprint} \colorbox{red!13.157894736842106}{\strut .} \colorbox{red!13.684210526315791}{\strut [SEP]} 
}}}
\end{CJK}

N: 
\begin{CJK}{UTF8}{gbsn}
{\setlength{\fboxsep}{0pt}\colorbox{white!0}{\parbox{0.35\textwidth}{
\colorbox{red!33.92857142857142}{\strut [CLS]} \colorbox{red!85.71428571428571}{\strut conclusions} \colorbox{red!28.57142857142857}{\strut are} \colorbox{red!42.857142857142854}{\strut provided} \colorbox{red!23.21428571428571}{\strut in} \colorbox{red!100.0}{\strut section} \colorbox{red!37.49999999999999}{\strut 4} \colorbox{red!21.428571428571427}{\strut .} \colorbox{red!37.49999999999999}{\strut [SEP]} 
}}}
\end{CJK}
\end{quote}
We also see that the dataset contains many relatively difficult instances, as we show in the following incorrectly classified examples. E.g., the model observes ``briefly discussed'' as an indicator that an instance is non-cite-worthy when it is in fact cite-worthy, and that ``described earlier'' and ``previous work'' signal that a sentence is cite-worthy when it is in fact labelled as non-cite-worthy.
\begin{quote}
\begin{CJK}{UTF8}{gbsn}
C: 
{\setlength{\fboxsep}{0pt}\colorbox{white!0}{\parbox{0.35\textwidth}{
\colorbox{red!26.190476190476193}{\strut [CLS]} \colorbox{red!76.19047619047619}{\strut some} \colorbox{red!71.42857142857143}{\strut approaches} \colorbox{red!23.80952380952381}{\strut for} \colorbox{red!20.476190476190478}{\strut the} \colorbox{red!57.142857142857146}{\strut solution} \colorbox{red!16.666666666666668}{\strut as} \colorbox{red!29.523809523809526}{\strut well} \colorbox{red!15.714285714285715}{\strut as} \colorbox{red!30.0}{\strut their} \colorbox{red!66.66666666666667}{\strut limitations} \colorbox{red!32.38095238095239}{\strut are} \colorbox{red!100.0}{\strut briefly} \colorbox{red!61.904761904761905}{\strut discussed} \colorbox{red!26.66666666666667}{\strut .} \colorbox{red!30.0}{\strut [SEP]} 
}}}
\end{CJK}

N:
\begin{CJK}{UTF8}{gbsn}
{\setlength{\fboxsep}{0pt}\colorbox{white!0}{\parbox{0.35\textwidth}{
\colorbox{red!28.358208955223876}{\strut [CLS]} \colorbox{red!34.32835820895522}{\strut this} \colorbox{red!43.28358208955223}{\strut simple} \colorbox{red!17.91044776119403}{\strut and} \colorbox{red!49.253731343283576}{\strut fast} \colorbox{red!49.253731343283576}{\strut technique} \colorbox{red!19.402985074626862}{\strut for} \colorbox{red!16.41791044776119}{\strut the} \colorbox{red!47.76119402985074}{\strut production} \colorbox{red!14.776119402985072}{\strut of} \colorbox{red!89.55223880597013}{\strut snps} \colorbox{red!40.298507462686565}{\strut was} \colorbox{red!68.65671641791045}{\strut described} \colorbox{red!99.99999999999999}{\strut earlier} \colorbox{red!41.7910447761194}{\strut in} \colorbox{red!70.14925373134326}{\strut our} \colorbox{red!86.56716417910447}{\strut previous} \colorbox{red!73.13432835820895}{\strut work} \colorbox{red!26.86567164179104}{\strut .} \colorbox{red!32.83582089552238}{\strut [SEP]} 
}}}
\end{CJK}
\end{quote}
We hypothesize that in such instances, context can help the most in disambiguating which sentences in a paragraph should be labelled as cite-worthy. Additionally, other information such as the section in which a sentence resides could help. E.g., to correctly label the fourth statement above as ``non-cite-worthy'', it may help to see that the last sentence of the paragraph is ``In our previously published work, it was reported that SNPs were joined together by the heat treatment, and this process led to increase in the sizes of SNPs which finally resulted in sharper XRD peaks'' which is a cite-worthy sentence. Additionally, it may help to know that it resides in the ``Discussion'' section of the paper.


\section{RQ3: Domain Evaluation}
\label{sec:domain-adaptation}
\begin{figure}[t]
  
  \centering
    \includegraphics[width=0.5\textwidth]{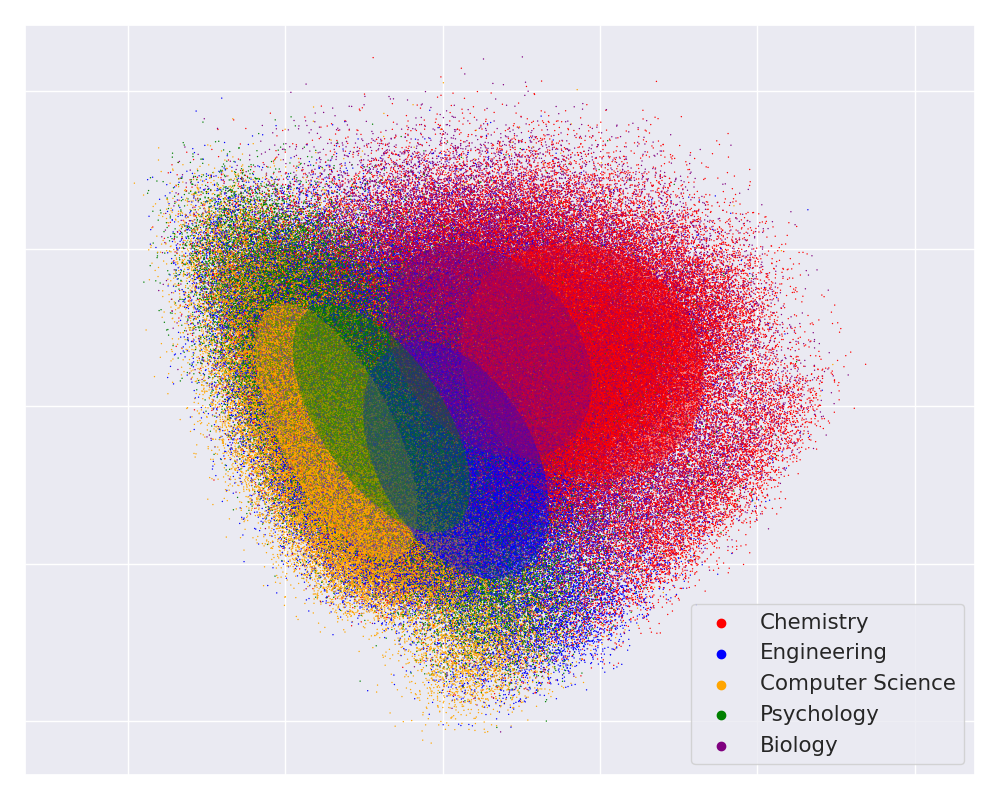}
    \caption{Visualizing the BERT embeddings for 5 of the 10 domains from \dataset~using the method by \citet{Aharoni2020UnsupervisedDC}. Clustering is performed using Gaussian Mixture Models.}
    \label{fig:domain-visualization}
\end{figure}

We next ask: how does domain affect learning to perform cite-worthiness? To answer this, we study the relationships between cite-worthiness data from different fields and how the Longformer-Ctx model performs in a cross-domain setup. For ease of analysis we limit the scope of fields to 5 of the 10 fields in the dataset: Chemistry, Engineering, Computer Science, Psychology, and Biology. 

First, we visualize the embedding space for data from each of these domains using the method of \citet{Aharoni2020UnsupervisedDC}. In this, the data is passed through BERT (specifically the base, uncased variant) and the output representations for each token in a sentence are average pooled. These representations are visualized in 2D space via PCA in \autoref{fig:domain-visualization}. It is clear that similar fields occupy closer space, with `engineering' and `computer science' sharing closer representations, as well as `biology' and `chemistry'. We perform clustering on this data using a Gaussian mixture model similarly to \citet{Aharoni2020UnsupervisedDC}, finding that domains form somewhat distinct clusters with a cluster purity of 57.61. This demonstrates that the data in different fields are drawn from different distributions, thus differences could exist in a model's ability to perform cite-worthiness detection on out of domain data.

To test this, we perform a cross-validation experiment using the 5 selected fields, training on one field and testing on another for all 25 combinations. The results for the 5x5 train/test setup using Longformer-Ctx are given in \autoref{tab:domain-adaptation-results}. 

Not surprisingly, the best performance for each split occurs when training on data from the same field. We also observe high variance in the maximum performance for each field ($\sigma$ = 3.32), and between different fields on the same test data, despite large pretrained Transformer models being relatively invariant across domains~\cite{wright2020transformer}. This suggests stark differences in how different fields employ citations. Additionally, we observe a strong (inverse) correlation between distance in the embedding space and performance on different domains, showing that using more similar data for training helps on out-of-domain performance~\cite{Aharoni2020UnsupervisedDC}. 


\begin{table}[t]
\def\arraystretch{1.5}
    \centering
    \fontsize{10}{10}\selectfont
    \begin{tabular}{l|c|c|c|c|c|}
    \multicolumn{1}{l}{\diagbox[innerwidth=1cm]{Train}{Test}} & \multicolumn{1}{c}{Ch} & \multicolumn{1}{c}{E} & \multicolumn{1}{c}{CS} & \multicolumn{1}{c}{P} & \multicolumn{1}{c}{B}\\
    \cline{2-6}
    Ch & \textbf{67.58}& 58.41& 56.86& 62.35& 68.23\\
    \cline{2-6}
    E & 66.62& \textbf{60.25}& 60.11& 64.02& 68.07\\
    \cline{2-6}
    CS & 65.05& 59.36& \textbf{61.99}& 63.85& 66.72\\
    \cline{2-6}
    P & 65.49& 58.03& 56.69& \textbf{65.10}& 68.27\\
    \cline{2-6}
    B & 66.59& 58.80& 58.22& 64.54& \textbf{69.12}\\
    \cline{2-6}
    \multicolumn{1}{c}{} \\
    \cline{2-6}
    $\sigma$ & 0.90& 0.78& 2.02& 0.92& 0.77\\
    \cline{2-6}
    $\rho$ & 0.87& 0.86& 0.76& 0.67& 0.79\\
    \cline{2-6}
    \end{tabular}
    \caption{F1 performance on different domain adaptation settings for the fields (Ch)emistry, (E)ngineering, (C)omputer (S)cience, (P)sychology, and (B)iology. 
    Out-of-domain tests use the entire set of data from that field, while in domain tests use 80\% of data for training, 10\% for validation, and 10\% for test. $\sigma$ is the standard deviation of performance of different train domains on the given test domain, and $\rho$ is Pearson correlation between performance and Euclidean distance from the train domain cluster to the test domain cluster. }
    \label{tab:domain-adaptation-results}
\end{table}

\section{RQ4: Cite-Worthiness for Transfer Learning}
\label{sec:transfer-learning}
\begin{table*}[t]
    \centering
    \fontsize{10}{10}\selectfont
    \begin{tabular}{l l l c c c c}
    \toprule 
    Dataset & Reference & Task & \multicolumn{1}{l}{Base} & \multicolumn{1}{l}{LM} &
    \multicolumn{1}{l}{Cite}& \multicolumn{1}{l}{LMCite}\\
    \midrule 
       BC5CDR & \citet{li2016biocreative} & NER & $89.84_{0.18}$& $\mathbf{90.03_{0.11}}$& $89.73_{0.25}$& $90.02_{0.79}$ \\
       JNLPBA & \citet{kim2004introduction} & NER & $77.02_{0.36}$& $77.13_{0.53}$& $76.97_{0.44}$& $\mathbf{77.15_{0.58}}$ \\
       NCBI-Disease & \citet{dougan2014ncbi}& NER & $\mathbf{88.79_{0.35}}$& $88.53_{0.58}$& $88.66_{0.57}$& $88.31_{0.43}$\\
       SciERC & \citet{luan-etal-2018-multi}& NER & $67.08_{0.50}$& $66.64_{0.47}$& $67.12_{0.46}$& $\mathbf{67.48_{0.45}}$ \\
    \midrule
       EBM-NLP & \citet{nye2018corpus} & PICO & $76.61_{0.21}$& $\mathbf{76.69_{0.28}}$& $76.55_{0.88}$& $76.41_{0.32}$ \\
    \midrule
       ChemProt & \citet{kringelum2016chemprot}& REL & $83.17_{0.43}$& $\mathbf{83.26_{0.90}}$& $82.70_{1.06}$& $83.16_{0.63}$\\
       SciERC &\citet{luan-etal-2018-multi} & REL & $80.21_{0.81}$& $\mathbf{80.68_{1.04}}$& $80.00_{1.73}$& $80.58_{0.96}$ \\
    \midrule
       ACL-ARC & \citet{jurgens2018measuring} & CLS & $71.82_{2.93}$& $70.95_{2.25}$& $\mathbf{73.68_{2.75}}$& $72.92_{3.76}$ \\
       SciCite & \citet{cohan2019structural} & CLS & $84.83_{0.65}$& $85.18_{0.47}$& $85.32_{0.16}$& $\mathbf{85.35_{0.29}}$ \\
       PaperField & \citet{beltagy2019scibert} & CLS & $65.48_{0.18}$& $\mathbf{65.57_{0.27}}$& $65.46_{0.24}$& $65.42_{0.48}$ \\
    \midrule
       Average & & & 78.386 & 78.466& 78.619 & \textbf{78.680} \\
    \bottomrule 

    \end{tabular}
    \caption{Performance on various downstream scientific document understanding tasks as presented by \citet{beltagy2019scibert}. The metrics used are the same as in their paper: NER is span-level F1, PICO is token level F1, relation extraction is macro-F1, and ChemProt is micro-F1. All runs are averaged across 5 seeds. Subscripts are the standard deviation for 5 runs.}
    \label{tab:downstream_task_results}
\end{table*}
The final question we ask is: to what extent is cite-worthiness detection transferable to downstream tasks in scientific document understanding? To answer this, we fine tune SciBERT on the task of cite-worthiness detection as well as masked language modeling (MLM) on \dataset, followed by fine-tuning on several document understanding tasks. We use SciBERT in order to have a direct comparison with previous work \cite{beltagy2019scibert}. The tasks we evaluate on come from~\citet{beltagy2019scibert} and are categorized as follows. 
\begin{itemize}[noitemsep]
    \item Named Entity Recognition (NER)/PICO: These tasks involve labelling the spans of different types of entities in a document.
    \item Relation Extraction (REL): This task involves labelling a sequence for the relationship between two entities.
    \item Text classification (CLS): Finally, we test on several text classification tasks (citation intent classification and paper field classification), where the goal is to classify a sentence into one or more categories. 
\end{itemize}
We compare five variants of pre-training and fine-tuning, given as follows.

\paragraph{Base} The original SciBERT model.

\paragraph{LM} SciBERT with MLM fine tuning on \dataset.

\paragraph{Cite} SciBERT fine-tuned for the task of cite-worthiness detection. The classifier is a pooling layer on top of the \texttt{[CLS]} representation of SciBERT, followed by a classification layer.

\paragraph{LMCite} SciBERT with MLM fine tuning and cite-worthiness detection. The two tasks are trained jointly i.e. on each batch of training, the model incurs a loss for both MLM and cite-worthiness detection which are summed together.

The results for all experiments are given in \autoref{tab:downstream_task_results}. Note that the reported results for SciBERT are on re-running the model locally for fair comparison. We first observe that incorporating our dataset into fine-tuning tends to improve model performance across all tasks to varying degrees, with the exception of NER on the NCBI-Disease corpus. The tasks where cite-worthiness as an objective has the most influence are the two citation intent classification tasks (ACL-ARC and SciCite). We see average improvements of 1.8 F1 points for the ACL-ARC dataset (including 2 points F1 improvement over the minumum and maximum model performance of SciBERT) and 0.5 F1 points on SciCite. 
The best average performance is from the model which incorporates both MLM and cite-worthiness as an objective, which we call \modelname.\footnote{We release two \modelname\ models available from the HuggingFace model hub: \texttt{copenlu/citebert} and \texttt{copenlu/citebert-cite-only}.}

For other tasks, fine-tuning the language model on \dataset~data tends to be sufficient for improving performance, though the margin of improvement tends to be minimal. This is in line with previous work reporting that language model fine-tuning on in-domain data leads to improvements on end-task fine-tuning \cite{DBLP:conf/acl/GururanganMSLBD20}. \dataset~is relatively small compared to the corpus on which SciBERT is originally trained (30.7M tokens for the train and dev splits on which we train versus 3.1B), so one could potentially see further improvements by incorporating more data or including cite-worthiness as an auxiliary task during language model pre-training. However, this is outside the scope of this work. 

\section{Conclusion}
In this work, we present an in-depth study into the problem of cite-worthiness detection in English. We rigorously curate \dataset, a high-quality dataset for cite-worthiness detection; present a paragraph-level contextualized model which improves by 5.31 F1 points on the task of cite-worthiness detection over the existing state-of-the-art; show that \dataset~is a good testbed for studying domain adaptation in scientific text; and show that in a transfer-learning setup one can achieve state of the art results on the task of citation intent classification using this data. In addition to studying cite-worthiness and transfer learning, \dataset~is suitable for use in downstream natural language understanding tasks. 
As we retain the S2ORC metadata with the data, one could potentially use the data to study joint cite-worthiness detection and citation recommendation. Additionally, one could explore other useful problems such as modeling different authors' writing styles and incorporating the author network as a signal. We hope that the data and accompanying fine-tuned models will be useful to the research community working on problems in the space of scientific language processing.

\section*{Acknowledgements}
$\begin{array}{l}\includegraphics[width=1cm]{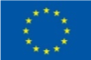} \end{array}$ The research documented in this paper has received funding from the European Union's Horizon 2020 research and innovation programme under the Marie Sk\l{}odowska-Curie grant agreement No 801199. 

\bibliography{anthology,acl2020}
\bibliographystyle{acl_natbib}

\clearpage

\appendix
\section{List of Permissible Section Titles}
\label{sec:premissible-section-titles}
\begin{itemize}[noitemsep]
    \item introduction
\item abstract
\item method
\item methods
\item results
\item discussion
\item discussions
\item conclusion
\item conclusions
\item results and discussion
\item related work
\item experimental results
\item literature review
\item experiments
\item background
\item methodology
\item conclusions and future work
\item related works
\item limitations
\item procedure
\item material and methods
\item discussion and conclusion
\item implementation
\item evaluation
\item performance evaluation
\item experiments and results
\item overview
\item experimental design
\item discussion and conclusions
\item results and discussions
\item motivation
\item proposed method
\item analysis
\item future work
\item results and analysis
\item implementation details
\end{itemize}
\section{List of Regular Expressions}
\label{sec:regexes}
Citation format regexes:
\begin{itemize}
    \item \texttt{\textbackslash[([0-9]+\textbackslash s*[,-;]*\textbackslash s*)*[0-9]+\textbackslash s*\textbackslash]}
    \item \texttt{\textbackslash(?[12][0-9]{3}[a-z]?\textbackslash s*\textbackslash)}
\end{itemize}
Hanging citation regex:
\texttt{\textbackslash s+\textbackslash(?(\textbackslash(\textbackslash s*\textbackslash)|like|reference|\\including|include|with|for instance|for example|see also|at|following|of|from|to|in|by|\\see|as|e\textbackslash .?g\textbackslash .?(,)?|viz(\textbackslash .)?(,)?)\textbackslash s*\\(,)*(-)*[\textbackslash)\textbackslash]]?\textbackslash s*[.?!]\textbackslash s*\$}

\section{Reproducibility}
\subsection{Computing Infrastructure}
All experiments were run on a shared cluster. Requested jobs consisted of 16GB of RAM and 4 Intel Xeon Silver 4110 CPUs. We used a single NVIDIA Titan X GPU with 12GB of RAM.

\subsection{Average Runtimes}
The average runtime performance of each model is given in \autoref{tab:runtimes}. Note that different runs may have been placed on different nodes within a shared cluster. 
\begin{table}
    \centering
    \fontsize{10}{10}\selectfont
    \begin{tabular}{l c c}
    \toprule 
     Setting & Time\\
    \midrule
        Logistic Regression & 00h01m43s\\
        Transformer & 02h55m13s\\
        BERT & 05h30m30s\\
        SciBERT (no weighting) & 09h22m00s\\
        SciBERT & 09h32m37s\\
        SciBERT + PU & 16h01m27s\\
        Longformer-Solo & 75h27m22s\\
        Longformer-Ctx & 19h16m07s\\
    \bottomrule 

    \end{tabular}
    \caption{Average runtimes for each model (runtimes are taken for the entire run of an experiment).}
    \label{tab:runtimes}
\end{table}

\subsection{Number of Parameters per Model}
The number of parameters in each model is given in \autoref{tab:num_params}.
\begin{table}
    \centering
    \fontsize{10}{10}\selectfont
    \begin{tabular}{l c c}
    \toprule 
     Method & \# Parameters\\
    \midrule
        Logistic Regression & 198,323\\
        Transformer & 9,789,042\\
        BERT & 109,484,290\\
        SciBERT & 109,920,514\\
        Longformer & 149,251,586\\
    \bottomrule 

    \end{tabular}
    \caption{Number of parameters in each model}
    \label{tab:num_params}
\end{table}

\subsection{Validation Performance}
The validation performance of each tested model is given in \autoref{tab:val_performance}.
\begin{table}
    \centering
    \fontsize{10}{10}\selectfont
    \begin{tabular}{l c c}
    \toprule 
     Method & F1\\
    \midrule
        Logistic Regression & -\\
        Transformer & 57.02\\
        BERT & 60.75\\
        SciBERT (no weighting) & 57.52\\
        SciBERT & 62.04\\
        SciBERT + PU & 61.43\\
        Longformer-Solo & 61.67\\
        Longformer-Ctx & 67.11\\
    \bottomrule 

    \end{tabular}
    \caption{Average validation performance for each of the models.}
    \label{tab:val_performance}
\end{table}

\subsection{Evaluation Metrics}
The primary evaluation metric used was F1 score.
We used the sklearn implementation of \texttt{precision\_recall\_fscore\_support} for F1 score, which can be found here: \url{https://scikit-learn.org/stable/modules/generated/sklearn.metrics.precision_recall_fscore_support.html}. Briefly:
\begin{equation*}
   p = \frac{tp}{tp + fp} 
\end{equation*}
\begin{equation*}
   r = \frac{tp}{tp + fn} 
\end{equation*}
\begin{equation*}
   F1 = \frac{2*p*r}{p+r} 
\end{equation*}
where $tp$ are true positives, $fp$ are false positives, and $fn$ are false negatives.

\subsection{Hyperparameters}
\label{sec:hyperparams}
\paragraph{Logistic Regression} We used a C value of 0.1151 for logistic regression.

\paragraph{Basic Transformer} The final hyperparameters for the basic Transformer model are: batch size: 64; number of epochs: 33; feed-forward dimension: 128; learning rate: 0.0001406; number of heads: 3; number of layers: 5; weight decay: 0.1; dropout probability: 0.4. We performed a Bayesian grid search over the following ranges of values, optimizing validation F1 performance: learning rate: $[0.000001, 0.001]$; batch size: $\{4, 8, 16, 32, 64, 128\}$; weight decay: $\{0.0, 0.0001, 0.001, 0.01, 0.1\}$; dropout probability: $\{0.0, 0.1, 0.2, 0.3, 0.4, 0.5\}$; number of epochs: $[2, 40]$; feed-forward dimension: $\{128, 256, 512, 1024, 2048\}$; number of heads: $\{1, 2, 3, 4, 5, 6, 10, 12\}$; number of layers: $[1, 12]$.

\paragraph{BERT} The final hyperparameters for BERT are: batch size: 8; number of epochs: 3; learning rate: 0.000008075; triangular learning rate warmup steps: 300; weight decay: 0.1; dropout probability: 0.1. We performed a Bayesian grid search over the following ranges of values, optimizer validation F1 performance: learning rate: $[0.0000001, 0.0001]$; triangular learning rate warmup steps: $\{0, 100, 200, 300, 400, 500, 1000, 1500, 2000,$ $2500, 5000\}$; batch size: $\{4, 8\}$; weight decay: $\{0.0, 0.0001, 0.001, 0.01, 0.1\}$; number of epochs: $[2, 40]$.

\paragraph{SciBERT} The final hyperparameters for SciBERT are: batch size: 4; number of epochs: 3; learning rate: 0.000001351; triangular learning rate warmup steps: 300; weight decay: 0.1; dropout probability: 0.1. We performed a Bayesian grid search over the following ranges of values, optimizer validation F1 performance: learning rate: $[0.0000001, 0.0001]$; triangular learning rate warmup steps: $\{0, 100, 200, 300, 400, 500, 1000, 1500, 2000,$ $2500, 5000\}$; batch size: $\{4, 8\}$; weight decay: $\{0.0, 0.0001, 0.001, 0.01, 0.1\}$; number of epochs: $[2, 40]$.

\paragraph{Longformer-Ctx} The final hyperparameters for Longformer-Ctx are: batch size: 4; number of epochs: 3; learning rate: 0.00001112; triangular learning rate warmup steps: 300; weight decay: 0.0; dropout probability: 0.1. We performed a Bayesian grid search over the following ranges of values, optimizer validation F1 performance: learning rate: $[0.0000001, 0.0001]$; triangular learning rate warmup steps: $\{0, 100, 200, 300, 400, 500, 1000, 1500, 2000,$ $2500, 5000\}$; batch size: $\{4, 8\}$; weight decay: $\{0.0, 0.0001, 0.001, 0.01, 0.1\}$; number of epochs: $[2, 6]$.

\subsection{Data}
\dataset~is constructed from the S2ORC dataset, which can be found here: \url{https://github.com/allenai/s2orc}. In particular, \dataset~is built using the \texttt{20200705v1} release of the data. A link to the \dataset~data can be found here: \url{https://github.com/copenlu/cite-worth}.
\end{document}